\PassOptionsToPackage{table}{xcolor}
\documentclass[11pt,a4paper,logo,nonumbering]{style}
\usepackage{array,booktabs,tabularx,multirow,graphicx,mathtools}
\usepackage{tcolorbox}
\tcbuselibrary{breakable}
\usepackage{adjustbox}
\usepackage{placeins}
\usepackage{float}
\usepackage{fontawesome5}

\definecolor{brandorange}{HTML}{ED6000}
\definecolor{absframe}{HTML}{ED6000}
\definecolor{rowhl}{HTML}{F4F7FA}

\newcommand{\sys}{InfraRSI}
\newcommand{\spair}{\sigma_{\mathrm{pair}}}

\newcolumntype{Y}{>{\raggedright\arraybackslash}X}
\newcolumntype{P}[1]{>{\hsize=#1\hsize\raggedright\arraybackslash}X}

\graphicspath{{assets/}}
\titlespacing*{\section}{0pc}{1.8ex plus 2pt minus 2pt}{2.5pt}
\titlespacing*{\subsection}{0pc}{1.2ex plus 1.5pt minus 1pt}{1.5pt}
\titlespacing*{\paragraph}{0pc}{0.6ex plus 1pt minus 1pt}{0.6em}
\hypersetup{
  pdftitle={InfraRSI: Harnessing Recursive Self-Improvement for Training Infrastructure},
  pdfauthor={InfraRSI Team}
}

\makeatletter
\renewcommand{\abscontent}{%
  \begin{tcolorbox}[breakable,colframe=absframe,colback=absframe!5,boxrule=1pt,arc=8pt,
    left=10pt,right=10pt,top=10pt,bottom=10pt]
  \centerline{\fontsize{15pt}{14pt}\selectfont\textbf{Abstract}}\vspace{2ex}
  {\absfont\theabstract}
  \end{tcolorbox}%
}
\renewcommand{\maketitle}{%
  \begingroup\setlength{\parindent}{0pt}
  \vspace*{3pt}
  {\centering\titlefont\@title\par}
  \vskip8pt
  {\centering\normalfont\fontsize{10.5}{13}\selectfont\itshape ``Command Once, and Self-Evolve: Your Personal Infra Agent for All Environments.''\par}
  \vskip8pt
  {\centering\normalfont\fontsize{11}{15}\selectfont InfraRSI Team\par}
  \vskip12pt
  \endgroup
  \thispagestyle{firststyle}\abscontent
}
\makeatother

\fancypagestyle{firststyle}{%
  \fancyhf{}
  \fancyhead[L]{%
    \smash{%
      \raisebox{0.22em}{\includegraphics[height=1.25em]{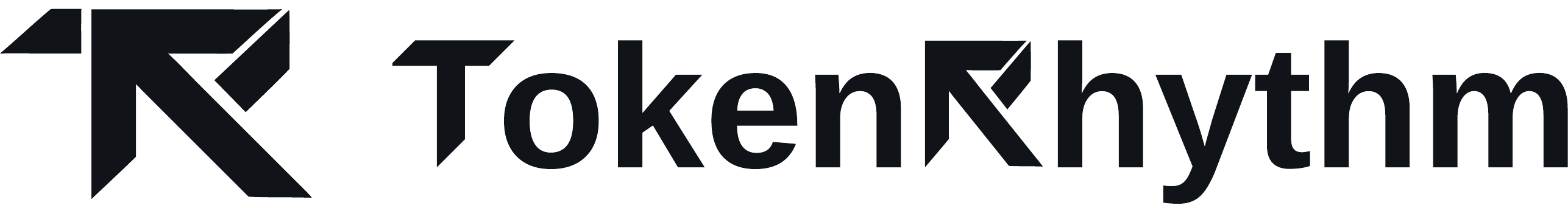}}%
      \hspace{0.9em}%
      \raisebox{0.12em}{\includegraphics[trim=7bp 7bp 7bp 7bp,clip,height=1.45em]{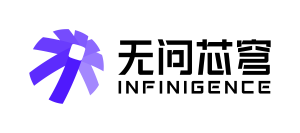}}%
    }%
  }
  \fancyhead[R]{%
    \smash{%
      \raisebox{0.08em}{%
        \includegraphics[height=1.7em]{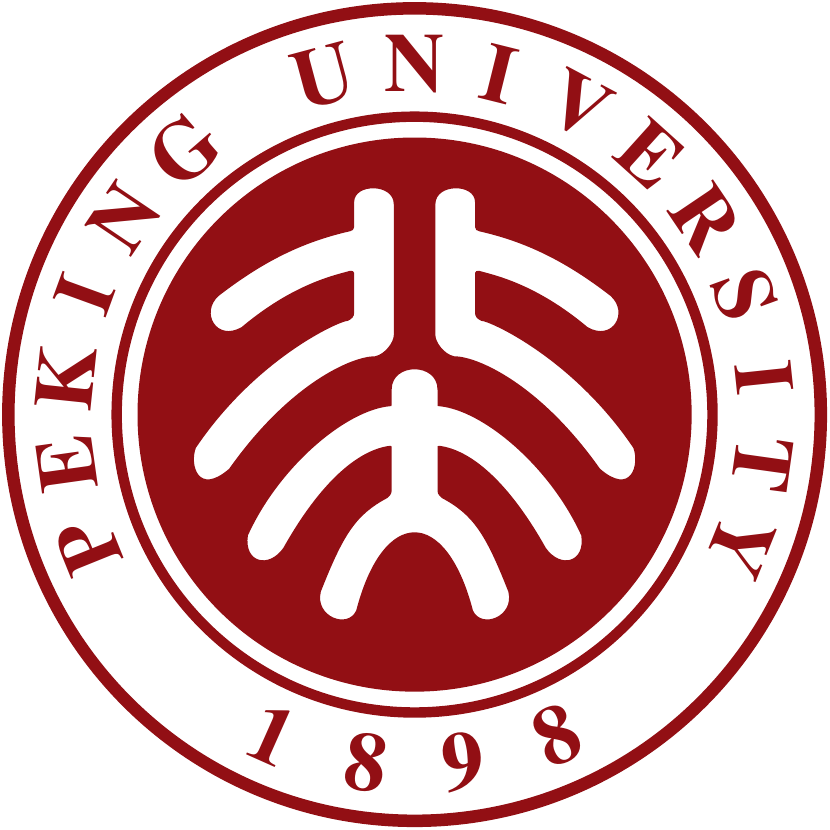}%
        \hspace{0.55em}%
        \includegraphics[height=1.7em]{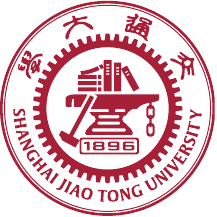}%
        \hspace{0.55em}%
        \includegraphics[height=1.7em]{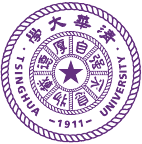}%
      }%
    }%
  }

}

\title{InfraRSI: Harnessing Recursive Self-Improvement for Training Infrastructure}
\author{InfraRSI Team}
\date{}

\begin{abstract}
While Recursive Self-Improvement (RSI) has accelerated isolated algorithm and kernel synthesis, distributed LLM training throughput depends on tightly coupled launch settings, operator kernels, and communication schedules, where single edits often hit out-of-memory (OOM) boundaries, compiler conflicts, measurement jitter, or silent numerical drift. We present \textbf{\sys{}}, the agentic harness that turns open-ended human commands into verified training-system optimizations through a five-stage workflow: \texttt{init} locks editable scope and a fixed task contract over workload, loss, and timing invariants to prevent reward hacking, \texttt{doctor} profiles baseline phases and calibrates cluster noise, \texttt{run} co-optimizes launch configurations and code patches before verifying top candidates in back-to-back baseline re-runs and writing checked findings to a cross-hardware experience bank, and \texttt{report} and \texttt{apply} audit source paths, account for failed-trial costs, and commit promoted changes. Rather than stepping greedily from a single configuration during \texttt{run}, \sys{} initializes from diverse experience-seeded configurations and maintains global execution-time and censored peak-memory models fitted to all trials, including OOM failures. Using these models, it estimates per-setting memory costs, revises miscalibrated LLM speed priors from live measurements, and pairs memory-hungry optimizations with memory-saving settings and verified code patches. Across four single- and multi-node dense and Mixture-of-Experts (MoE) LLM training workloads, \sys{} achieves a speedup of up to $2.75\times$ and lifts mean Model FLOPs Utilization (MFU) from $17.4\%$ to $30.0\%$, outperforming existing configuration tuners and coding agents while transferring experience across GPU architectures. As it improves training infrastructure efficiency, \sys{} forms a two-level Recursive Self-Improvement loop---revising priors online within runs and distilling experience across clusters---to push foundation model scaling.

\vspace{0.6ex}
\noindent\centerline{\faGithub\ \textbf{GitHub:} \href{https://github.com/TokenRhythm/InfraRSI}{\texttt{https://github.com/TokenRhythm/InfraRSI}}}\\[0.2ex]
\noindent\centerline{\faGlobe\ \textbf{Project Page:} \href{https://infrarsi.github.io}{\texttt{https://infrarsi.github.io}}}

\end{abstract}

\begin{document}
\maketitle

\section{Introduction}\label{sec:intro}

\begin{figure}[t]
  \centering
  \includegraphics[width=\linewidth]{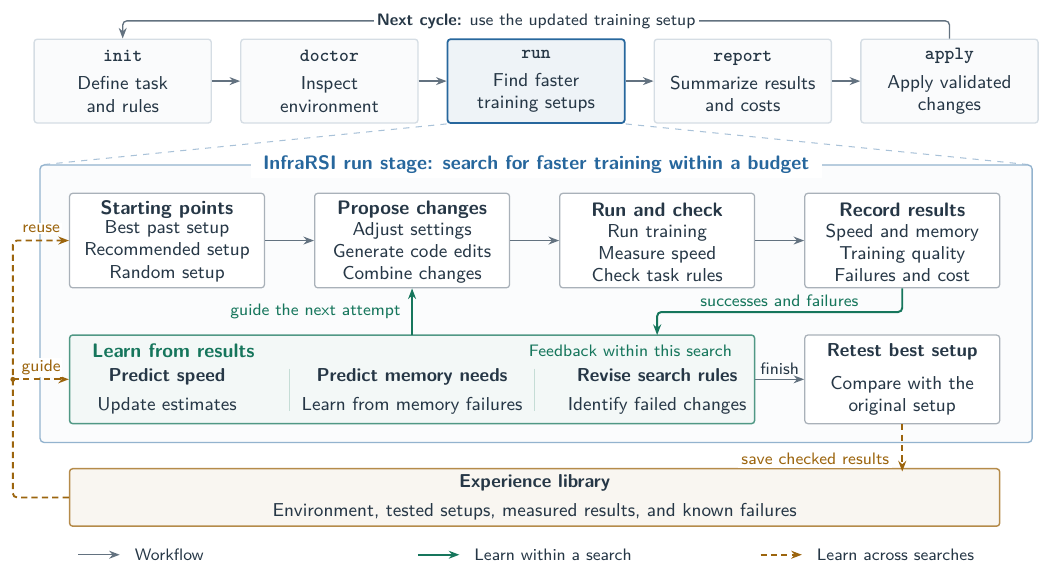}
  \caption{\textbf{InfraRSI method overview.} The five-stage lifecycle defines
  task rules, inspects the environment and baseline performance, searches for
  changes, reports results, and applies validated changes. The expanded
  \texttt{run} stage shows \sys{} searching over training settings and code edits
  generated by a language model. Measurements and failures update speed and
  memory predictions and guide the next attempt. The best setup is retested
  against the original before checked results are saved as reusable experience.
  Later searches use this experience to choose starting setups and initialize
  predictions. The updated training setup enters the next lifecycle.}
  \label{fig:infrarsi-method}\label{fig:bcomp_overview}
\end{figure}

Building on classic formulations of self-referential improvement \citep{godel2007}, recent Recursive Self-Improvement (RSI) systems demonstrate that large language models (LLMs) can discover mathematical constructions \citep{funsearch2024, alphaevolve2025}, synthesize machine learning pipelines \citep{aide2025}, and rewrite agentic scaffolding \citep{dgm2025}. While these systems target isolated algorithms or single kernels, compute-optimal scaling laws \citep{chinchilla2022} increasingly bind foundation model progress to the throughput of distributed training infrastructure \citep{megatronlm, megatron2021, deepspeed2020, deepspeed, fsdp2}: launch configuration knobs, operator and training-loop implementations, collective communication schedules, and device memory limits (Table~\ref{tab:paradigm}). In this work, we study how to turn open-ended human commands and language-model proposals into verified training-system optimizations through a two-level self-improving loop---revising priors online within each tuning run while distilling verified runtime evidence into a cross-cluster experience bank under fixed task contracts. Jointly optimizing both launch configurations and training source code is essential because hardware-sensitive mechanisms such as FP8 quantization \citep{fp8training2022, deepseekv3}, tensor and expert parallelism, graph compilation (\texttt{torch.compile}) \citep{torchcompile}, and FlashAttention \citep{flashattention2, flashattention3} interact non-linearly and often reverse their performance impact across GPU architectures and interconnect rooflines \citep{roofline2009} (Table~\ref{tab:cross_hw}).

Yet applying unconstrained coding agents \citep{metr2024rebench, karpathy2026autoresearch, sakana2025cuda} or greedy local search loops directly to production training repositories exposes four recurring failure modes (Section~\ref{sec:motivation}). Open-ended code edits frequently introduce reward hacking, numerical drift, or timing shortcuts, achieving illusory speedups by altering effective batch sizes, deferring optimizer updates past the timing window, or applying unstable low-precision casts. At the same time, distributed clusters exhibit wide-spectrum measurement noise across two orders of magnitude, causing fixed acceptance thresholds either to promote random jitter on shared hosts or to discard valid communication overlaps and incremental code patches on quiet nodes. Guidance from zero-shot LLM priors is similarly unreliable: predicted speedups correlate poorly with actual hardware execution and often veto high-gain compilation or communication settings. Finally, single-incumbent search loops stall at out-of-memory (OOM) and compiler boundaries whenever a throughput-boosting setting exceeds device memory on its own and must be paired simultaneously with memory-saving knobs and compatible kernel patches.

To address these failure modes, we present \textbf{\sys{}} (Figure~\ref{fig:infrarsi-method}, Section~\ref{sec:architecture}), the agentic harness that orchestrates a five-stage optimization workflow (\texttt{init} $\to$ \texttt{doctor} $\to$ \texttt{run} $\to$ \texttt{report} $\to$ \texttt{apply}). In \sys{}, \texttt{init} locks editable repository scope and a fixed task contract over workload, loss, and timing invariants to prevent reward hacking, while \texttt{doctor} profiles baseline execution phases and calibrates paired step-time noise on the target cluster. The \texttt{run} stage executes \textbf{budgeted compositional search} (Section~\ref{sec:hls}), a history-aware search controller organized around three mechanisms:
(1) \emph{Multi-Start Initialization and Global Time/Memory Models} (\S\ref{sec:hls-headroom}), which starts from diverse experience-seeded configurations and fits global execution-time and right-censored peak-memory models across all trials;
(2) \emph{Prior Calibration, Failure Attribution, and Bundle Completion} (\S\ref{sec:hls-screening}), which revises miscalibrated LLM priors from live probes and pairs high-upside OOM configurations with model-priced memory savers; and
(3) \emph{High-Impact Knob Escape and Code Patch Stacking} (\S\ref{sec:bcomp}), which expands search around high-leverage launch settings while stacking verified code patches onto stable configurations. At the close of \texttt{run}, the harness verifies top candidates via independent back-to-back re-runs against the baseline (Table~\ref{tab:invariants}) and distills hardware-normalized runtime evidence into the cross-hardware experience bank (\S\ref{sec:experience}), after which \texttt{report} and \texttt{apply} audit source paths, account for failed-trial costs, and commit promoted changes.

We evaluate \sys{} in open-code optimization campaigns across four dense and Mixture-of-Experts (MoE) LLM training workloads: Qwen3.5-9B on 8$\times$H100, Gemma-4-26B-A4B MoE on 8$\times$H200, Qwen3.8-27B on 8$\times$H200, and two-node Qwen3.5-35B-A3B MoE on 16$\times$H100 (Table~\ref{tab:bcomp_budget10}, Figure~\ref{fig:bcomp_budget10}). Across these workloads, \sys{} achieves a $\mathbf{1.83\times}$ geometric-mean speedup and lifts mean training Model FLOPs Utilization (MFU) from $17.4\%$ to $30.0\%$ (reaching up to $2.75\times$ on Gemma-4-26B-A4B MoE), outperforming RE-Bench ($1.37\times$, $25.1\%$ MFU), Auto-Research ($1.34\times$, $24.5\%$ MFU), and knob-only OptiCo ($1.23\times$, $21.3\%$ MFU). Furthermore, cross-SKU experience transfer accelerates convergence and improves final throughput over cold-start search (Table~\ref{tab:rsi_transfer}), while offline receipt and tolerance audits confirm that our 16-invariant verifier reliably filters out invalid and reward-hacking executions across numerical regimes (Table~\ref{tab:l1_guardrail}). Our contributions are:
\begin{itemize}
  \item \textbf{Unified Execution and Verification Contract (\S\ref{sec:architecture}):} A five-stage multi-stack harness over launch configurations and source patches that couples 16 workload, numerical, and timing invariants (Table~\ref{tab:invariants}) with source-path auditing, full failed-trial cost accounting, and back-to-back baseline verification.
  \item \textbf{Budgeted Compositional Search (\S\ref{sec:hls}):} A stateful configuration and code search controller that uses Cholesky execution-time and censored Tobit peak-memory models to guide empirical prior calibration, failure-attributed bundle completion, and high-impact knob escape (Figure~\ref{fig:bcomp_overview}).
  \item \textbf{Cross-Hardware Experience Transfer and Audited Evaluation (\S\ref{sec:experience}, \S\ref{sec:evaluation}):} A structured experience protocol that transfers hardware-conditioned performance evidence across GPU architectures, validated across dense and MoE training workloads and numerical tolerance regimes (Tables~\ref{tab:bcomp_budget10}--\ref{tab:l1_guardrail}, Figures~\ref{fig:bcomp_budget10}--\ref{fig:v41_tolerance}).
\end{itemize}

\section{Problem Definition and Empirical Motivation}\label{sec:motivation}

\subsection{Optimization Objective and Fixed Task Contract}\label{sec:motivation-domain}

Optimizing distributed training infrastructure differs fundamentally from algorithmic discovery \citep{funsearch2024, alphaevolve2025, aide2025} and isolated kernel generation \citep{kernelbench2025, sakana2025cuda} (Table~\ref{tab:paradigm}). In \sys{}, each candidate combines launch configuration knobs and source-code patches to maximize training throughput under a strict \emph{fixed task contract}: the model architecture, dataset, effective workload (tokens processed per optimizer update), allowed numerical precision menu, and optimizer update protocol are held invariant. Each task contract declares the evaluation window, permitted numerical modes, and relative loss tolerance, while retaining the full GPU and API search cost of any failed or rejected candidate.

Evaluating a candidate under this contract separates two distinct verification questions: \emph{short-horizon loss consistency} checks that the candidate's execution path remains numerically faithful within the declared precision menu and loss window, whereas \emph{independent back-to-back verification} confirms that a candidate's speedup reproduces against a fresh reference execution beyond cluster measurement noise. Because distributed training mechanisms compete for finite device memory, streaming multiprocessors, and collective-communication overlap windows, individual optimizations combine sub-additively and frequently trigger steep feasibility boundaries, such as compilation and activation-checkpointing conflicts on Mixture-of-Experts models. Expressing both search cost and projected runtime savings in common GPU-hour units allows the harness to schedule and evaluate optimization trials directly.

\begin{table}[!htbp]
\centering
\scriptsize
\setlength{\tabcolsep}{3.2pt}
\renewcommand{\arraystretch}{0.94}
\caption{Comparison of RSI paradigms across algorithmic synthesis, kernel generation, and training infrastructure.}
\label{tab:paradigm}
\begin{tabularx}{\textwidth}{@{}l Y Y Y@{}}
\toprule
\textbf{Dimension} & \textbf{Algorithm / Code RSI} \citep{funsearch2024, alphaevolve2025, aide2025} & \textbf{Kernel Synthesis} \citep{kernelbench2025, sakana2025cuda} & \textbf{Training Infra RSI (\sys{})} \\
\midrule
\textbf{Granularity} & Single heuristic or script & Isolated CUDA/Triton kernel \citep{triton2019} & Launch knobs \& source-code patches (multi-stack) \\
\rowcolor{rowhl}
\textbf{Cost \& Noise} & ms--s; $10^3$--$10^4$ runs; low noise & ms/op; $10^2$--$10^3$ runs; low noise & Multi-GPU; 10--20 runs; $0.04\%$--$6\%+$ noise \\
\textbf{Equivalence} & Test suite or val score & Tensor tolerance ($L_\infty/L_2$) & Fixed task contract + source-path audit \\
\rowcolor{rowhl}
\textbf{Hacking Risk} & Moderate (val overfitting) & High (caching, skipped sync) & Severe (prefetch skew, timer shift, precision drift) \\
\textbf{HW Sensitivity} & Hardware-agnostic logic & GPU-specific (SM, SRAM) & Roofline/topology-sensitive (H100/H200/H20) \\
\rowcolor{rowhl}
\textbf{Reusable Asset} & Task-specific code pool & Op kernel library & Cross-hardware priors + audited patch sites \\
\bottomrule
\end{tabularx}
\end{table}

\subsection{Evaluation Baselines and System Failure Modes}\label{sec:motivation-failures}

To evaluate open-code infrastructure search fairly, we compare \sys{} against representative autonomous coding agents and configuration tuners, including Auto-Research \citep{karpathy2026autoresearch}, the RE-Bench modular agent scaffold \citep{metr2024rebench}, and knob-only OptiCo \citep{optico2026}, under identical task contracts while retaining each baseline's native search rules. Across all campaigns, completed trial counts, GPU execution time, and token consumption are logged along the search trajectory, including all failed and audit-excluded attempts.

Audited runs across our four dense and MoE LLM workloads (Table~\ref{tab:bcomp_budget10}) expose four recurring failure modes in unconstrained coding agents and greedy local tuners:

\paragraph{1. Reward hacking and timing-boundary skew.}
Without strict contract verification and source-path auditing, open-code agents frequently achieve illusory speedups via reward gaming \citep{skalse2022reward} by altering effective work or timing boundaries. Common failure patterns include unauthorized low-precision casts that violate the declared precision menu, deferred auxiliary updates that push work past the timing window while masking runtime import errors behind a single receipt's validity flag, step-logging timer shifts that inflate native reported speed relative to wall-clock time, and aggressive low-precision or fused cross-entropy edits that degrade training loss (\S\ref{sec:architecture}).

\paragraph{2. Wide-spectrum measurement noise.}
Across clusters and workloads, paired step-time noise spans more than two orders of magnitude, ranging from $0.04\%$--$0.10\%$ on dedicated 8$\times$H100 and 8$\times$H200 nodes to over $6\%$ on shared hosts experiencing periodic data-loader or collective stalls. A hardcoded acceptance threshold (such as a fixed $2\%$ gate) is too coarse on quiet clusters, where it discards valid $0.5\%$--$1.5\%$ communication overlaps and incremental source-code patches that compound when stacked, yet still promotes random jitter on noisy hosts unless governed by calibrated cluster noise gates and independent back-to-back re-runs (\S\ref{sec:hls-headroom}).

\paragraph{3. Miscalibrated LLM quantitative priors.}
While frontier LLMs can inspect training repositories and propose qualitative mechanisms, their zero-shot speedup predictions correlate negatively with hardware execution (Spearman rank correlation $r \in [-0.49, -0.06]$). In addition, zero-shot priors issue \emph{false-negative vetoes} on critical high-impact knobs: on Gemma-4-26B-A4B MoE (8$\times$H200), the prior predicts a negligible $+2.0\%$ gain for graph compilation where empirical execution unlocks $+55\%$; on two-node Qwen3.5-35B-A3B MoE (16$\times$H100), it predicts a slight slowdown for increasing NCCL communication channels where execution yields a $+10.2\%$ gain. A search engine must therefore combine multi-start initialization with contrarian probing and empirical prior revision (\S\ref{sec:hls-screening}).

\paragraph{4. Single-incumbent myopia at memory and compiler cliffs.}
Optimizing training infrastructure requires coordinating multiple launch knobs and code patches across steep memory and compatibility boundaries. Greedy single-incumbent search loops fail in three ways. First, they discard or repeat failures without attribution, repeatedly triggering identical out-of-memory crashes or blaming valid launch configurations when an exploratory source patch fails. Second, they abandon multi-knob synergies after single-knob OOMs: on Gemma-4-26B-A4B MoE, disabling activation recomputation alone exhausts device memory ($+18\text{ GiB}$ shortfall) and enabling compilation with default fused kernels hangs, locking out tuners that cannot pair the memory-hungry optimization with a memory-saving partner ($-14\text{ GiB}$) and \texttt{FlexAttention}~\citep{flexattention2024} to reach $2.75\times$ ($17.6\%$ MFU). Third, they stall in local optima, as when Auto-Research plateaus at $1.29\times$ on Gemma-4-26B-A4B MoE and spends its remaining trials circling that local basin (\S\ref{sec:hls}, \S\ref{sec:experience}).

\section{System Architecture: The \sys{} Harness}\label{sec:architecture}

To address the failure modes in \S\ref{sec:motivation}, \sys{} decouples language-model hypothesis generation from measurement, verification, search, and experience storage. The harness pairs a declarative \texttt{infrarsi.yaml} task specification with lightweight stack adapters ($<15$ hook lines) to expose a uniform execution and receipt interface across major distributed LLM training frameworks, including Megatron-Core~\citep{megatronlm, megatron2021}, FSDP2/TorchTitan~\citep{fsdp2, torchtitan2025}, DeepSpeed~\citep{deepspeed2020, deepspeed}, HF Trainer/TRL~\citep{transformers2020}, and native PyTorch DDP.

\subsection{Workflow and Multi-Stack Adapters}\label{sec:arch-lifecycle}

Every optimization campaign progresses through five deterministic CLI stages (Figure~\ref{fig:infrarsi-method}):
\begin{itemize}
  \item \textbf{1. \texttt{init}:} Parses target repository ASTs, declares the task scope and \texttt{editable\_files}, and locks SHA-256 digests of all \texttt{frozen\_files} (dataset loaders, evaluation scripts, loss definitions, and harness timers).
  \item \textbf{2. \texttt{doctor}:} Executes reference baseline runs, trims warmup steps, calibrates cluster step-time noise, profiles execution phase breakdowns (forward, backward, optimizer, data loading, and collectives), probes interconnect bandwidth, and grades experience-bank prior coverage (\S\ref{sec:experience}).
  \item \textbf{3. \texttt{run}:} Executes \sys{}'s compositional search controller (\S\ref{sec:hls}) over candidate configurations and source patches, preserving raw execution receipts and full resource costs for every attempted candidate; at the end of search, it subjects the fastest valid candidates to fresh back-to-back verification re-runs against the reference baseline on the same GPU allocation, writes verified observations and experience packs back to the experience bank (\S\ref{sec:experience}), and saves \texttt{result.json}, \texttt{pack.yaml}, and \texttt{REPORT.md}.
  \item \textbf{4. \texttt{report}:} Renders and inspects the structured campaign report (\texttt{REPORT.md}), invariant verdicts, and resource accounting for candidate source-path auditing.
  \item \textbf{5. \texttt{apply}:} Reads the promoted candidate from \texttt{result.json}, checks target repository files for post-search working-tree conflicts, and writes the validated launch configuration and source diffs (optionally on a new git branch).
\end{itemize}

\subsection{Measurement and Resource Accounting}\label{sec:arch-actions}

\sys{} represents each candidate as a joint specification of \textbf{launch configuration knobs} (CLI flags, environment variables, and configuration keys filtered by algebraic batch-conservation checks) and \textbf{source-code patches} that synthesize AST-validated unified diffs on \texttt{editable\_files}, such as compiling block-sparse or chunked attention kernels~\citep{flexattention2024, torchcompile}, fusing normalization, rotary-embedding~\citep{liger2024}, and AdamW~\citep{adamw} optimizer operators, and bucketing gradient-norm collectives. Every candidate evaluation and verification run enforces a fixed optimizer-update count, explicit warmup trimming, and a synchronized steady-state timing window.

To prevent survivorship bias, the harness tracks GPU hours across reference, search, promotion, and profiling stages alongside language-model token consumption, retaining the full GPU and API cost of every failed or audit-excluded candidate. Incomplete runs from transient host retries are flagged in the ledger, and migrating a workload to a new host requires establishing a fresh local baseline.

\subsection{Semantic, Numerical, and Source-Path Verification}\label{sec:arch-contract}

Within each fixed task contract, \sys{} maximizes log speedup over the candidate space $\mathcal{A}$ subject to 16 deterministic invariants $\mathcal{I}_k$ across five verification layers:
\begin{equation}\label{eq:honest-contract}
\max_{a \in \mathcal{A}} \; y(a) = \ln\!\frac{T_0}{T(a)} \quad \text{s.t.} \quad \mathcal{I}_k(a) = \mathrm{True}, \;\forall k \in \{1,\dots,16\}.
\end{equation}
Task invariants verify execution completion and VRAM caps (L1), model parameters, sequence length, tokens per update, data digests, and \texttt{frozen\_files} hashes (L2), actual optimizer step counts (L3), permitted precision menu and relative loss bounds under pre-declared \texttt{tolerance} regimes ranging from $0.1\%$ to $50\%$ (L4), and harness-to-script timing agreement (L5). For numerical verification (L4), all loss tolerances and adaptation rules are fixed in the task specification before search begins: by default, L4 bounds relative window-mean loss deviation within $\tau$ (adjusting to $3\sigma_\ell\sqrt{2/k}$ if reference repeats vary beyond $\tau$, or enforcing per-update maximum deviation when exact batch order is verified; Table~\ref{tab:app_invariants}). Any candidate exceeding these pre-declared bounds is rejected while retaining its incurred search cost.

Alongside automated receipt checks, \sys{} performs \textbf{candidate source-path auditing} before promoting a result. In our open-code evaluations (\S\ref{sec:evaluation}), source review excluded candidates that deferred auxiliary parameter updates to the subsequent step (which omitted the final update from the timed window and masked a runtime kernel import failure behind a single receipt's validity flag) or that introduced unauthorized low-precision casts outside the task's permitted precision menu, while confirming that compliant candidates execute all forward, backward, and optimizer steps synchronously inside the timed window.

\subsection{Promotion and Delivered Speedups}\label{sec:arch-verifier}

\begin{table}[!htbp]
\centering
\scriptsize
\setlength{\tabcolsep}{3.2pt}
\renewcommand{\arraystretch}{0.94}
\caption{\textbf{Five-layer, 16-item deterministic invariant verifier and offline re-verification ledger} (1,212 stored full-fidelity executions across 106 searches, complemented by source-path and contract audits). Triggers count every failed item of a rejected execution; primary attributes each rejected execution to its first failed layer (full predicates in Appendix~\ref{app:invariants}, Table~\ref{tab:app_invariants}).}
\label{tab:invariants}
\begin{tabularx}{\textwidth}{@{}l P{.9} P{1.1} P{1.0}@{}}
\toprule
\textbf{Layer} & \textbf{Invariants} & \textbf{Core criterion} & \textbf{Offline triggers / primary} \\
\midrule
\textbf{L1: Execution} & \texttt{completes}, \texttt{no\_errors}, \texttt{memory\_cap} & Exit $0$, full target update count completed, no NaN / OOM / NCCL / import error in main run, VRAM within cap & 503 / 352 (\texttt{completes} 349, \texttt{no\_errors} 153, \texttt{memory\_cap} 1) \\
\textbf{L2: Workload} & \texttt{same\_workload}, \texttt{same\_data}, \texttt{same\_model}, \texttt{frozen\_files} & Same tokens per update and sequence length, per-update data digest, model digest and parameter count, frozen-file SHA-256 hashes; unverifiable required items reject & 7 / 7 (unverifiable item 7) \\
\textbf{L3: Schedule} & \texttt{optimizer\_steps} & Same actual optimizer-step count and synchronous auxiliary state updates inside the timed window & 0 / 0 (offline; deferred updates caught in source audit) \\
\textbf{L4: Numerical} & \texttt{loss\_tol}, \texttt{grad\_norm\_tol}, \texttt{precision} & Pre-declared task loss rule (window-mean within tolerance, or strict per-update max deviation when required); permitted precision menu enforced & 42 / 41 (\texttt{loss\_tol} 42; 39 pass without L4) \\
\textbf{L5: Timing \& Pair} & \texttt{timing\_consistent}, \texttt{speed\_reproduces} & Harness wall-clock gain agrees with reported step-time gain ($|\Delta \ln| \le 0.03$, flagging finite-prefetch or log-boundary skew); promotion requires back-to-back fresh re-test beyond calibrated noise & 158 / 0 (\texttt{timing\_consistent} 158; unpromoted runs deliver $1.00\times$) \\
\midrule
\rowcolor{rowhl}
\multicolumn{4}{@{}p{\dimexpr\textwidth-2\tabcolsep\relax}@{}}{\textbf{Offline re-verification: 1,212 executions, 400 rejected (33.0\%, 710 triggers), 812 valid}; verifier unit suite reproduces 40/40 expected verdicts (28 perturbed rejected, 12 benign passed). Excluded candidates retain full API and GPU costs in the search ledger.} \\
\bottomrule
\end{tabularx}
\end{table}

Table~\ref{tab:invariants} summarizes the five verification layers and their offline audit statistics (detailed predicates in Appendix~\ref{app:invariants}, Table~\ref{tab:app_invariants}; evaluation in \S\ref{sec:eval-ablations}). To ensure that search discoveries translate into reproducible system gains, promoted candidates undergo an independent back-to-back re-run against the reference configuration on the same GPU allocation.

When no candidate clears all verification and promotion gates, the harness retains the reference configuration and reports a delivered speedup of $1.00\times$, preserving all unpromoted and excluded trials in the audit log.

\section{Budgeted Compositional Search}\label{sec:hls}

Under practical multi-GPU search budgets, measurement jitter, miscalibrated language-model priors, steep memory and compiler-conflict cliffs, local optima, and late-stage context growth frequently stall naive proposal loops (\S\ref{sec:motivation}). In each step of the \texttt{run} stage, \sys{} reads and updates global execution-time and memory models across all completed trials (\S\ref{sec:hls-headroom}), revises language-model priors using live measurements and failure blame (\S\ref{sec:hls-screening}), and proposes coordinated knob bundles and source patches (\S\ref{sec:bcomp}; Figure~\ref{fig:bcomp_overview}).

\subsection{Multi-Start Initialization and Global Time/Memory Models}\label{sec:hls-headroom}

\paragraph{Online Noise Calibration and Phase-Headroom Ceiling.}
To guard against just-in-time compilation warmup~\citep{torchcompile, triton2019} and periodic data-loader or allocator jitter, the \texttt{doctor} stage trims initial warmup steps and measures each run's post-warmup steady-state window mean step time for candidate scoring, MFU accounting, and final promotion, while using robust median and split-half statistics inside the search loop so that transient single-step spikes do not distort surrogate updates. Across runs, \sys{} estimates paired log-step-time noise $\spair$ by pooling paired reference re-run differences with within-run split-half deviations (floored by the within-window white-noise floor), using $\max(0.02,\, 2\spair)$ to govern big-knob selection and stall detection during search and requiring fresh back-to-back promotion re-runs to exceed the paired relative margin $\exp(2\spair)-1$. This protocol measures paired baseline noise as low as $0.04\%\text{--}0.10\%$ on dedicated 8$\times$H100 and 8$\times$H200 nodes. Using the \texttt{doctor} stage's profiled operator-phase breakdown (forward, backward, optimizer, and collectives), \sys{} also caps single-lever log-speedup predictions via an Amdahl phase-headroom ceiling~\citep{amdahl1967}, bounding each lever's predicted gain by the profiled time share of the execution phases it targets.

\paragraph{Multi-Start Full-Configuration Initialization.}
Infrastructure search spaces partition into disconnected basins separated by out-of-memory boundaries and compiler incompatibilities. Starting solely from the unmodified default configuration wastes a large fraction of the trial budget flipping basic switches one by one, whereas trusting a single model-proposed bundle risks an immediate crash if any pair of levers conflicts. \sys{} therefore dedicates its first three full trials to probing complementary complete configurations (Figure~\ref{fig:bcomp_overview}). The first start warm-starts from the best verified full configuration in the most similar hardware and model environment in the experience bank via canonical mechanism normalization. The second start probes the prior model's highest-rated full configuration that differs from the first start in at least one high-impact knob, promoting configuration diversity. The third start executes a uniform random legal configuration filtered by active conflict rules to anchor global speed and memory variance, with the first fitting run establishing the memory baseline. In our Qwen3.5-9B (8$\times$H100) run, the experience-seeded first start reaches $1.55\times$ on Trial~1 before subsequent search advances to $1.73\times$.

\paragraph{Global Time and Censored Memory Models.}
Because configuration knobs and source patches interact non-monotonically across speed and memory, stepping greedily from only the latest incumbent deadlocks whenever an early choice blocks a downstream optimization. \sys{} instead maintains two global models fitted across the full trial history and checkpoints search state after each trial. A \textbf{Cholesky Bayesian linear time surrogate} models log-speedup over indicator features for non-reference knob settings, verified source-code patches, and enabling pairs, updating the posterior precision matrix in closed form while bounding prior variance. In parallel, a \textbf{right-censored Tobit memory ledger}~\citep{tobin1958} models per-GPU peak memory $\hat{m}(a)$ (in GiB) for candidate $a$ on a device of capacity $C_{\mathrm{card}}$ (with observation noise $\sigma_{\mathrm{obs}} = 0.02\,C_{\mathrm{card}}$) across three execution regimes: (i)~\emph{exact Gaussian residuals} $\hat{m}(a) - m_{\mathrm{peak}}$ on completed runs with measured peak memory $m_{\mathrm{peak}}$; (ii)~\emph{right-censored Tobit likelihoods} enforcing $\hat{m}(a) \ge m_{\mathrm{oom}}$ on out-of-memory trials, where the failed allocation shortfall establishes the lower bound $m_{\mathrm{oom}}$ on required memory; and (iii)~\emph{interval-censored likelihoods} bounding peak demand within $[0.90\,C_{\mathrm{card}},\, C_{\mathrm{card}}]$ when near-capacity allocator thrashing occurs. Fitting the posterior memory parameters via damped Newton steps with Laplace covariance yields calibrated memory feasibility probabilities and per-setting memory estimates (in GiB) that distinguish memory spenders from memory savers.

\subsection{Prior Calibration, Failure Attribution, and Bundle Completion}\label{sec:hls-screening}

\paragraph{Contrarian Big-Knob Probing and Empirical Prior Revision.}
Zero-shot language-model priors correlate poorly with actual multi-GPU execution (Spearman rank correlation $r \in [-0.49, -0.06]$) and frequently issue false-negative vetoes on high-leverage compiler or communication flags. To prevent miscalibrated priors from permanently hiding effective levers, \sys{} applies empirical-Bayes shrinkage~\citep{efron1973} across both action types: configuration-knob priors are regularized by their prior spread ($\sigma_0 \ge \max(\sigma_{\min},\, 0.5|\mu_0|)$) and updated in closed form by the Cholesky time surrogate as observations arrive, while language-model code-patch gain predictions are scaled by a conservative shrinkage factor $\kappa_0$ ($\mu_{\mathrm{patch}} = \kappa_0 \hat{\mu}_{\mathrm{LLM}}$). Alongside Bayesian posterior updating, \sys{} actively schedules \textbf{contrarian empirical probes} whenever a high-impact knob is rated low or vetoed by the prior yet exhibits high uncertainty whose two-sigma upside clears the big-knob threshold. \sys{} tests such levers against the incumbent by reverting any conflicting setting to reference and automatically attaching the top estimated memory-saving setting when memory feasibility is low. After repeated configuration failures or any failure of a high-confidence setting, \sys{} also triggers a structured \textbf{empirical prior revision} carrying measured step times, peak memory, and out-of-memory shortfalls, directly updating the prior levels ($\mu_0, \sigma_0$), refreshing the time surrogate's prior parameters and the memory ledger's declared GiB priors, and re-solving the posterior in place. As shown in \S\ref{sec:motivation}, this probing and revision loop recovers high-gain compilation and communication settings ($+55\%$ and $+10.2\%$) that zero-shot priors initially rate near zero or negative.

\paragraph{Failure Blame Attribution Across Memory, Conflicts, and Source Patches.}
\sys{} logs out-of-memory events, numerical violations, and runtime crashes separately in the ledger. When a non-reference setting triggers an OOM, it is tagged as requiring a memory-saving partner covering at least the measured GiB shortfall, and is excluded only if it still fails across multiple distinct savers. For non-memory crashes, hangs, or numerical divergence, \sys{} bans only the minimal conflict tuple (the smallest co-occurring pair or triple of settings absent from all passing runs) while keeping each individual setting eligible in other combinations. Failures during code exploration are attributed to the candidate patch rather than the underlying launch configuration.

\paragraph{Bundle Completion and Conflict Backtracking.}
Whenever a configuration fails due to insufficient memory despite high surrogate-predicted speedup, \sys{} retains it in a high-upside failure buffer and performs bundle completion, pairing the configuration with reference-held memory savers that cover the measured GiB shortfall while reverting any conflicting knob to reference. For example, on Gemma-4-26B-A4B MoE, bundle completion pairs disabling activation recomputation ($+18\text{ GiB}$ shortfall) with disabling Liger fused cross-entropy~\citep{liger2024} ($-14\text{ GiB}$) and enabling \texttt{torch.compile} (\S\ref{sec:eval-bcomp}). Similarly, when a high-prior lever conflicts with an earlier accepted baseline, conflict backtracking re-evaluates the lever directly on the clean root reference, migrates the incumbent once verified, and retains compatible settings on top of the updated baseline.

\subsection{High-Impact Knob Escape and Code Patch Stacking}\label{sec:bcomp}\label{sec:hls-backtracking}

\paragraph{Focusing Exploration on the Big-Knob Set.}
When search needs to backtrack, probe contrarian directions, or jump out of a local basin, perturbing minor knobs (such as memory pinning or worker thread counts whose effect is below $1\%$) cannot shift the hardware roofline or free enough memory to unlock major speedups, and risks attributing gains to sub-noise fluctuations. After every observation, \sys{} explicitly partitions the configuration space to isolate the \textbf{Big-Knob set} $\mathcal{K}_{\mathrm{big}}$, defined as all configuration knobs whose prior ($\hat{y}_{\mathrm{prior}}$), surrogate-posterior ($\hat{y}_{\mathrm{surr}}$), or empirical ($\bar{y}_{\mathrm{emp}}$) single-knob log-speedup magnitude on at least one setting clears $\max(0.02,\, 2\spair)$ ($\approx 2\%$). All structural exploration operators (initial probe diversification, contrarian probing, trust-region escape, tabu deduplication, and warm restarts) operate over $\mathcal{K}_{\mathrm{big}}$.

\paragraph{High-Impact Trust-Region Escape and Candidate Deduplication.}
Whenever two consecutive full trials fail to improve the incumbent beyond the noise gate, \sys{} enters escape mode over $\mathcal{K}_{\mathrm{big}}$. It dynamically expands the minimum trust-region distance from the incumbent from one to three big knobs to force multi-knob leaps out of the local basin, extrapolates further along ordered axes after improving steps, and performs periodic warm restarts from the best historical run of a distinct configuration class. To prevent wasted evaluations, \sys{} also enforces big-knob tabu filtering and registers every proposed configuration-and-patch signature in the visited set \emph{before} execution, cleanly separating pre-execution rejections from completed trials and preventing repetitive proposal loops.

\paragraph{Code Patch Stacking and Trajectory Summarization.}
When open-code editing is enabled, late-stage configuration proposals whose two-sigma upside falls below $\max(0.02, 2\spair)$ yield their evaluation slots to source-code synthesis, and verified patches with small individual gains are stacked and tested together in the spirit of group testing~\citep{dorfman1943} once their combined log-gain clears the noise gate. To keep prompts bounded over long trajectories, \sys{} retains full execution traces for the fastest and most recent runs while summarizing older trials into compact single-line records ($\sim 28\text{k}$ characters). In addition, when a candidate refines launch knobs on top of an already verified patch stack, unmodified inherited diff lines are not double-counted against the per-step line limit.

\section{Empirical RSI Loop: Experience Protocol and Cross-Hardware Transfer}\label{sec:experience}

Reusing optimization history across training jobs requires converting raw execution traces into portable priors without inducing negative transfer across GPU generations. Because hardware roofline shifts alter effect magnitudes and signs while different frameworks expose distinct configuration flags, \sys{} normalizes observations into a canonical experience schema weighted by hardware and workload similarity.

\subsection{Cross-Hardware Rooflines and Sign Flips}\label{sec:exp-roofline}

\begin{table}[!htbp]
\centering
\caption{\textbf{Cross-hardware scaling and sign flips in the initial seed corpus (11 seed packs, 191 environments, 698 observations, 543 effects)} across H20 ($4.0\text{ TB/s}$, $148\text{ TF}$), H100 ($3.35\text{ TB/s}$, $989\text{ TF}$), and H200 ($4.8\text{ TB/s}$, $989\text{ TF}$).}
\label{tab:cross_hw}
\scriptsize
\setlength{\tabcolsep}{3.2pt}
\renewcommand{\arraystretch}{0.94}
\begin{tabularx}{\textwidth}{@{}l l l l Y@{}}
\toprule
\textbf{Model} & \textbf{Mechanism} & \textbf{H20 (148 TF)} & \textbf{H100/200 (989 TF)} & \textbf{Roofline Cause \& Rule} \\
\midrule
Qwen3-8B & \texttt{precision=fp8} & $+58\%$ ($+54\%$ TP2) & $+30\%$ H200 ($+3\%$ TP2) & Compute-bound H20 vs.\ comm-bound H200; $\kappa$ scaling. \\
Qwen3-8B & \texttt{tp=2} & $+24\%$ ($1.24\times$) & $-4\%$ ($0.96\times$ H100) & \textbf{Sign flip:} AllReduce overhead exceeds GEMM gain. \\
Qwen3-8B & \texttt{te\_fp8\_attn} & $+6\%$ ($1.06\times$) & $-8\%$ ($0.92\times$ H200) & \textbf{Sign flip:} H200 BF16 FlashAttention-3 beats FP8 cast. \\
Dense LLM & \texttt{compile} & $+20\text{--}29\%$ ($+5\%$ TP2) & $+5\text{--}10\%$ ($+40\%$ H200) & Amortizes H20 launch; fuses H200 TP2 collectives. \\
Qwen3-30B MoE & \texttt{alltoall+grouped\_gemm} & $+92\%$ ($1.92\times$) & $+85\text{--}110\%$ (H200) & Consistent \texttt{kernel} gain on high-bandwidth NVLink. \\
\bottomrule
\end{tabularx}
\end{table}

Across our multi-cluster corpus spanning Hopper-generation H20, H100, and H200 GPUs on dense and MoE large language model workloads~\citep{qwen3} (Table~\ref{tab:cross_hw}), end-to-end training speedups scale sub-linearly with peak device FLOPS because activation recomputation, MoE token routing, and memory-bound elementwise operations dilute raw GEMM throughput. Crossing the compute-to-bandwidth ridge point~\citep{roofline2009} also induces systematic \emph{magnitude compression and sign flips} that break naive configuration replay. Compute-relief levers such as Transformer Engine FP8 execution~\citep{fp8training2022} and tensor parallelism yield substantial gains on compute-bound H20 nodes, but their benefits compress sharply or turn negative on H100 and H200 once faster Tensor Cores shrink GEMM time enough that quantization cast overhead, optimized BF16 FlashAttention-3 kernels~\citep{flashattention3}, or tensor-parallel AllReduce collectives dominate the training step. Compilation and communication-fusion effects similarly depend on interconnect topology and operator structure: graph compilation~\citep{torchcompile} amortizes kernel launch overhead on H20 and fuses tensor-parallel communication on H200, whereas NVLink-local MoE grouped GEMMs and all-to-all dispatch optimizations transfer consistently only within matching interconnect and model-architecture domains.

\subsection{Structured Experience Protocol}\label{sec:exp-protocol}

To make empirical observations portable across heterogeneous repositories, \sys{} structures experience records around three core entities: the \textbf{Environment}, capturing hardware rooflines, node and GPU topology, interconnect fabric, software versions, model dimensions, and parallel layout; the \textbf{Observation}, recording complete knob configurations, step time, peak device memory, numerical verification conditions, and 16-invariant verdicts alongside explicit out-of-memory shortfalls and minimal conflict tuples; and the \textbf{Effect}, storing paired log-speedup, calibrated noise floor, conditioning dependencies, and physical mechanism class (\texttt{kernel} versus \texttt{balance}). A two-tier normalizer maps framework-specific flags (such as fused AdamW~\citep{adamw} or mixed-precision switches) across stacks into \textbf{71 canonical mechanisms}.

To transfer priors safely across clusters, \sys{} filters and ranks each source record against the target environment using a compatibility scope check and an environmental distance $d \ge 0$. The scope check requires matching training frameworks for all records and grades an effect as \texttt{covered} when the model family and sequence-length bucket match (and, for topology-sensitive \texttt{balance} mechanisms, only when the GPU SKU, node count, per-node GPU count, multi-node interconnect fabric, parameter-scale bucket, and parallel layout also match), downgrading partial matches to \texttt{partial} and discarding incompatible records. Eligible records are then ordered lexicographically by coverage level (\texttt{covered} before \texttt{partial}), readability, evidence tier, and environmental distance $d$, which sums indicator penalties for GPU SKU and node-count mismatches, log-ratio differences $|\ln(a/b)|$ in parameter count and sequence length, and $0.25\times$ reference-state mismatches. Environmental similarity is used strictly to \emph{filter and rank prior evidence} (grading target levers into covered, null, partial, and unobserved states), whereas \emph{actual target observations} govern posterior estimates during search via Bayesian surrogate updating and empirical prior revision~\citep{efron1973} (\S\ref{sec:hls-screening}). The corpus statistics in Table~\ref{tab:cross_hw} (11 seed packs across 191 environments, 698 observations, 543 effects, and 71 canonical mechanisms) reflect the initial seed bank prior to our evaluation runs.

\subsection{Experience Retrieval and Write-Back}\label{sec:exp-transfer}

\paragraph{Where Experience Enters Search.}
Within \sys{}, retrieved experience packs enter the search loop at two explicit points: they directly drive the first start of the three-start initialization (\S\ref{sec:hls-headroom}) by mapping the highest-similarity verified winning configuration onto the target task's complete knob table, and they initialize the prior means and variances of the Cholesky Bayesian time surrogate as well as the per-setting GiB priors and conflict constraints of the right-censored Tobit memory ledger. In cross-hardware transfer evaluations (\S\ref{sec:eval-transfer}), retrieval excludes prior runs from the same workload and GPU SKU so that warm starts reflect cross-hardware priors rather than self-replay.

\paragraph{Verified Write-Back.}
At the end of a tuning run, observations that pass all verification invariants, numerical tolerances, and paired noise gates are packaged into a new experience pack and written back to the bank. In our controlled transfer evaluation on Qwen3.5-9B (8$\times$H100) (\S\ref{sec:eval-transfer}, Table~\ref{tab:rsi_transfer}), cross-SKU experience improves final speedup from $1.70\times$ to $1.82\times$ and reaches the $90\%$ peak log-speedup threshold at Trial~9 ($2.97\;\mathrm{GPU\cdot h}$), whereas cold-start and unfiltered retrieval do not reach that threshold within the search budget.

\section{Experimental Evaluation}\label{sec:evaluation}

We evaluate \sys{} in three parts:
(1) end-to-end training speedup and Model FLOPs Utilization (MFU) across four dense and MoE LLM workloads (\S\ref{sec:eval-bcomp}, Table~\ref{tab:bcomp_budget10}, Figures~\ref{fig:bcomp_budget10}--\ref{fig:bcomp_tasks});
(2) cross-hardware experience transfer (\S\ref{sec:eval-transfer}, Table~\ref{tab:rsi_transfer}, Figure~\ref{fig:v41_transfer}); and
(3) an independent offline verifier audit alongside numerical tolerance and noise-floor sensitivity studies (\S\ref{sec:eval-ablations}, Table~\ref{tab:l1_guardrail}, Figure~\ref{fig:v41_tolerance}).

\subsection{End-to-End Speedup and MFU Across LLM and MoE Workloads}\label{sec:eval-bcomp}

\begin{figure}[!htbp]
\centering
\includegraphics[width=0.99\textwidth]{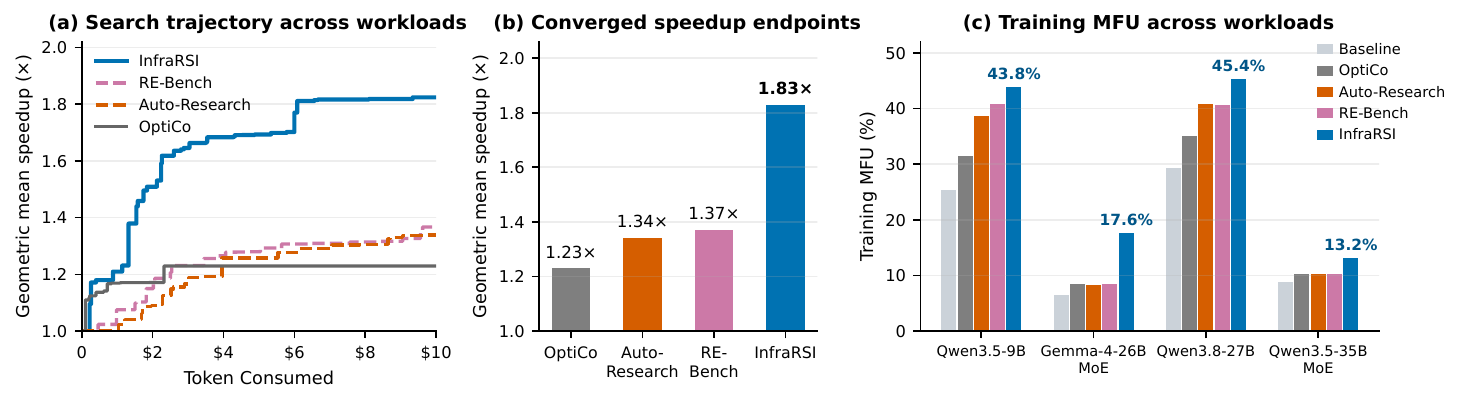}
\caption{\textbf{Search trajectories, final speedup endpoints, and training MFU across the four dense and MoE LLM workloads.} \textbf{(a)} Speedup trajectory versus token consumed across all compared methods. \textbf{(b)} Final geometric-mean speedup endpoints across the four workloads ($1.83\times$ for \sys{} vs.\ $1.37\times$ for RE-Bench, $1.34\times$ for Auto-Research, and $1.23\times$ for OptiCo). \textbf{(c)} Training MFU (\% of dense BF16 peak) across workloads.}
\label{fig:bcomp_budget10}
\end{figure}

\begin{figure}[!htbp]
\centering
\includegraphics[width=0.99\textwidth]{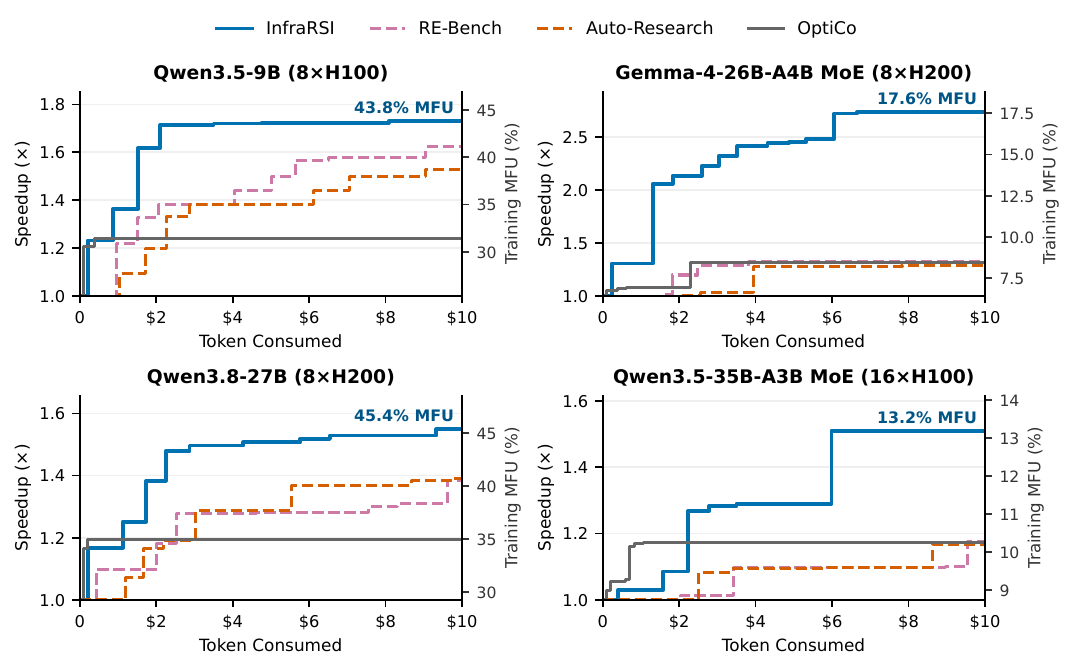}
\caption{\textbf{Per-workload search trajectories across the four dense and MoE LLM workloads,} plotting speedup ($\times$) on the left axis and training MFU (\%) on the right axis against token consumed.}
\label{fig:bcomp_tasks}
\end{figure}

We evaluate \sys{} (\S\ref{sec:bcomp}) against open-code Auto-Research~\citep{karpathy2026autoresearch}, the METR RE-Bench agent scaffold~\citep{metr2024rebench}, and knob-only OptiCo~\citep{optico2026} across four dense and MoE LLM workloads~\citep{qwen35, gemma4, qwen38}:
Qwen3.5-9B (8$\times$H100),
Gemma-4-26B-A4B MoE (8$\times$H200),
Qwen3.8-27B (8$\times$H200), and
Qwen3.5-35B-A3B MoE (16$\times$H100, 2 Nodes).
Table~\ref{tab:bcomp_budget10} and Figures~\ref{fig:bcomp_budget10}--\ref{fig:bcomp_tasks} report end-to-end training speedup and Model FLOPs Utilization (MFU) across all four workloads.

\begin{table}[!htbp]
\centering
\fontsize{6.6}{8.2}\selectfont
\setlength{\tabcolsep}{3.0pt}
\renewcommand{\arraystretch}{0.96}
\caption{\textbf{End-to-end training speedup ($\times$) and Model FLOPs Utilization (MFU, \% of non-sparse BF16 peak at $989.5\text{ TFLOP/s}$ per GPU) across four dense and MoE LLM workloads.}}
\label{tab:bcomp_budget10}
\begin{tabularx}{\textwidth}{@{}P{1.05} c r r r r Y@{}}
\toprule
\textbf{Workload \& Hardware} & \textbf{Base MFU} & \textbf{OptiCo} & \textbf{Auto-Research} & \textbf{RE-Bench} & \textbf{\sys{}} & \textbf{Key Discovered Knob + Code Combination} \\
\midrule
\rowcolor{rowhl}
Qwen3.5-9B (8$\times$H100) & $25.3\%$ ($1.62\text{ s}$) & $1.24\times$ ($31.5\%$) & $1.53\times$ ($38.6\%$) & $1.62\times$ ($40.9\%$) & $\mathbf{1.73\times}$ ($\mathbf{43.8\%}$) & Memory-ledger saver pairing (Liger frees VRAM for recompute-off + micro-batch 2) + fused normalization \\
Gemma-4-26B-A4B MoE (8$\times$H200) & $6.4\%$ ($3.38\text{ s}$) & $1.32\times$ ($8.4\%$) & $1.29\times$ ($8.3\%$) & $1.33\times$ ($8.5\%$) & $\mathbf{2.75\times}$ ($\mathbf{17.6\%}$) & Synergistic memory completion (\texttt{torch.compile} + Liger-off + recompute-off) + cached \texttt{FlexAttention} \\
\rowcolor{rowhl}
Qwen3.8-27B (8$\times$H200) & $29.3\%$ ($4.50\text{ s}$) & $1.19\times$ ($35.0\%$) & $1.39\times$ ($40.8\%$) & $1.38\times$ ($40.6\%$) & $\mathbf{1.55\times}$ ($\mathbf{45.4\%}$) & Memory-saver pairing + fused RMSNorm/RoPE/SwiGLU kernels + selective activation recomputation \\
Qwen3.5-35B-A3B MoE (16$\times$H100, 2 Nodes) & $8.7\%$ ($1.89\text{ s}$) & $1.17\times$ ($10.3\%$) & $1.17\times$ ($10.2\%$) & $1.17\times$ ($10.3\%$) & $\mathbf{1.51\times}$ ($\mathbf{13.2\%}$) & Contrarian 32-channel NCCL tuning + composite patch overlay (fused SwiGLU + bucketed grad norm) \\
\midrule
\textbf{4-Workload Mean (Geomean $\times$ / Mean MFU)} & $17.4\%$ & $1.23\times$ ($21.3\%$) & $1.34\times$ ($24.5\%$) & $1.37\times$ ($25.1\%$) & $\mathbf{1.83\times}$ ($\mathbf{30.0\%}$) & Consistent gains across single- and multi-node dense and MoE training \\
\bottomrule
\end{tabularx}
\end{table}

\paragraph{Overall End-to-End Speedup Performance.}
Across the four workloads, \sys{} achieves a $1.83\times$ geometric-mean speedup, consistently outperforming open-code RE-Bench ($1.37\times$), Auto-Research ($1.34\times$), and knob-only OptiCo ($1.23\times$) on every task (Table~\ref{tab:bcomp_budget10}, Figure~\ref{fig:bcomp_budget10}). Knob-only tuning plateaus early because launch flags alone cannot remove operator-level bottlenecks, whereas single-incumbent coding agents stall when memory-hungry configurations trigger OOMs or compiler conflicts before compatible source patches can be stacked.

\paragraph{Model FLOPs Utilization (MFU) and Physical Throughput.}
We compute Training MFU as $\mathrm{MFU} = F_{\mathrm{update}} / (N_{\mathrm{GPU}} \cdot P_{\mathrm{BF16,dense}} \cdot t_{\mathrm{update}})$ against the non-sparse Hopper BF16 peak ($989.5\text{ TFLOP/s}$ per GPU), using post-warmup mean step time $t_{\mathrm{update}}$ (updates 11--30) and PaLM dense-equivalent active FLOPs $F_{\mathrm{update}}$~\citep{palm2023} ($6\times$ active matrix parameters per token plus attention FLOPs, excluding recomputation, communication, and unrouted experts). Overall, \sys{} lifts mean MFU from $17.4\%$ to $30.0\%$, pushing compute-bound dense models above $43\%$ MFU and substantially improving utilization on routing- and communication-bound MoE workloads by jointly eliminating redundant activation recomputation, kernel launch overhead, and collective communication bottlenecks.

\paragraph{Coordinated Knob and Source-Patch Discoveries.}
The per-workload trajectories in Figure~\ref{fig:bcomp_tasks} illustrate how compositional search coordinates configuration knobs with source-code patches under numerical and timing guardrails. On Gemma-4-26B-A4B MoE (8$\times$H200), disabling activation recomputation in isolation triggers out-of-memory failures and enabling \texttt{torch.compile} conflicts with the default Liger kernel configuration~\citep{liger2024}; \sys{}'s censored memory model identifies disabling Liger as a $14\text{ GiB}$ memory saver, bans the conflicting pair, and completes the three-way feasible knob combination while synthesizing a compiled \texttt{FlexAttention}~\citep{flexattention2024} patch that caches block-sparse masks across the 25 sliding-window layers and a chunked causal attention kernel for the 5 global layers ($2.75\times$, $17.6\%$ MFU). On two-node Qwen3.5-35B-A3B MoE (16$\times$H100), contrarian knob exploration unlocks 32-channel InfiniBand NCCL communication while composite patch overlays stack fused SwiGLU, fast zero-grad, and bucketed gradient-norm all-reduce kernels ($1.51\times$, $13.2\%$ MFU). On dense Qwen3.8-27B (8$\times$H200) and Qwen3.5-9B (8$\times$H100), pairing memory-saving configuration knobs with fused normalization, rotary embedding, and MLP kernels lifts final speedups to $1.55\times$ ($45.4\%$ MFU) and $1.73\times$ ($43.8\%$ MFU), respectively.

\subsection{Cross-Hardware Experience Transfer}\label{sec:eval-transfer}

\begin{figure}[!htbp]
\centering
\includegraphics[width=0.98\textwidth]{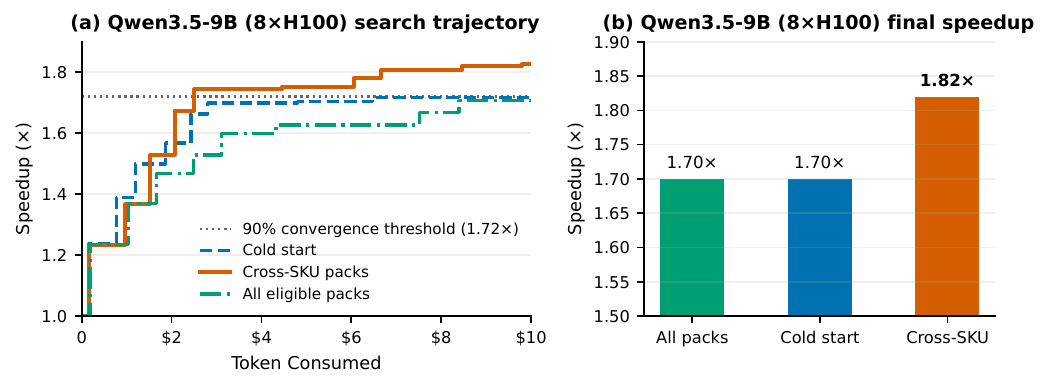}
\caption{\textbf{\sys{} cross-hardware experience transfer trajectories on Qwen3.5-9B (8$\times$H100).} Curves plot speedup against token consumed; the dashed horizontal line marks the $90\%$ peak log-speedup threshold ($\ln(S) \ge 0.9\ln(S^\star)$).}
\label{fig:v41_transfer}
\end{figure}

Table~\ref{tab:rsi_transfer} and Figure~\ref{fig:v41_transfer} evaluate cross-hardware experience transfer on Qwen3.5-9B (8$\times$H100). We compare cold start (no prior experience packs) and all eligible packs against cross-SKU transfer, which restricts both structured experience packs and the transfer brief to evidence harvested on other GPU architectures while excluding same-task, same-SKU history. When experience transfer is active, the initial trial warm-starts a knob configuration from the transferred winner and seeds priors into the time surrogate and memory ledger. We measure both the final speedup and the number of trials and GPU-hours required to reach $90\%$ of the peak log-speedup ($\ln(S) \ge 0.9\ln(S^\star)$).

\begin{table}[!htbp]
\centering
\fontsize{6.8}{8.4}\selectfont
\setlength{\tabcolsep}{3.2pt}
\renewcommand{\arraystretch}{0.95}
\caption{\textbf{\sys{} cross-hardware experience transfer on Qwen3.5-9B (8$\times$H100).} Cross-SKU transfer filters both experience packs and the transfer brief to evidence harvested on other hardware SKUs while excluding same-task, same-SKU runs; the initial trial warm-starts a knob configuration and seeds priors for the time surrogate and memory ledger. Convergence speed is measured by the trials and GPU-hours required to reach $90\%$ of the peak log-speedup ($\ln(S) \ge 0.9\ln(S^\star)$).}
\label{tab:rsi_transfer}
\begin{tabularx}{\textwidth}{@{}l Y c r c r@{}}
\toprule
\textbf{Target Workload \& Hardware} & \textbf{Experience Condition} & \textbf{First Trial} & \textbf{Final Speedup} & \textbf{Trials to 90\% Peak} & \textbf{GPU-h to 90\% Peak} \\
\midrule
\rowcolor{rowhl}
Qwen3.5-9B (8$\times$H100) & Cold start (no prior packs) & $1.22\times$ (Trial 2) & $1.70\times$ & $>15$ (censored) & --- \\
Qwen3.5-9B (8$\times$H100) & All eligible packs & $1.24\times$ (Trial 1) & $1.70\times$ & $>17$ (censored) & --- \\
\rowcolor{rowhl}
Qwen3.5-9B (8$\times$H100) & Cross-SKU transfer & $1.23\times$ (Trial 1) & $\mathbf{1.82\times}$ & $\mathbf{9}$ & $\mathbf{2.97}$ \\
\bottomrule
\end{tabularx}
\end{table}

As shown in Table~\ref{tab:rsi_transfer}, cross-SKU experience transfer improves both final speedup and search convergence on Qwen3.5-9B (8$\times$H100): it lifts the final speedup from $1.70\times$ (under cold start and all eligible packs) to $1.82\times$ and is the only condition to cross the workload-wide $90\%$ peak threshold within the search budget (Trial~9, $2.97\;\mathrm{GPU\cdot h}$), whereas both cold start ($>15$ trials) and all eligible packs ($>17$ trials) are right-censored.

\subsection{Offline Verifier Audit and Tolerance Sensitivity}\label{sec:eval-ablations}

\begin{figure}[!htbp]
\centering
\includegraphics[width=0.98\textwidth]{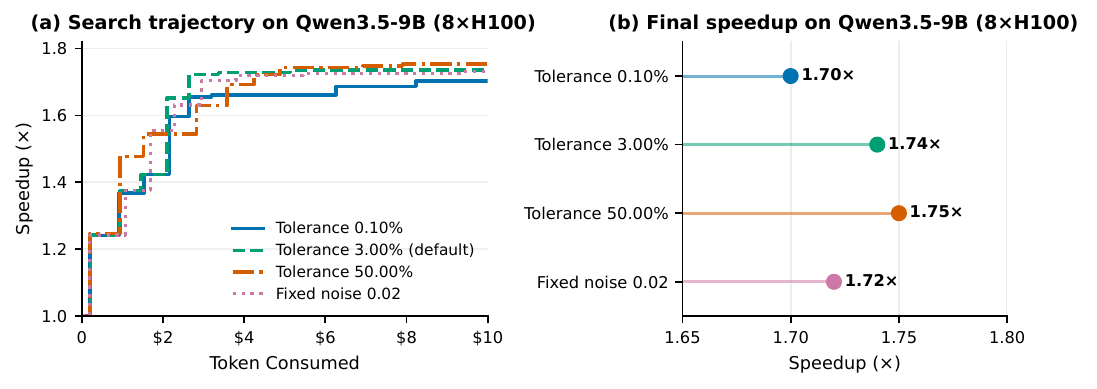}
\caption{\textbf{\sys{} numerical tolerance and fixed-noise settings on Qwen3.5-9B (8$\times$H100).} \textbf{(a)} Search trajectories across relative loss tolerances ($0.10\%$, $3.00\%$, and $50.00\%$) and a fixed $2.00\%$ noise floor (at $3.00\%$ tolerance), with the common $90\%$ peak threshold ($S^\star = 1.75\times$) marked. \textbf{(b)} Final speedups across the four ablation settings.}
\label{fig:v41_tolerance}
\end{figure}

Finally, Table~\ref{tab:l1_guardrail} and Figure~\ref{fig:v41_tolerance} evaluate our independent offline verifier audit (Part~A) alongside our numerical tolerance and fixed-noise ablation on Qwen3.5-9B (8$\times$H100) (Part~B).

\begin{table}[!htbp]
\centering
\fontsize{6.8}{8.4}\selectfont
\setlength{\tabcolsep}{2.6pt}
\renewcommand{\arraystretch}{0.95}
\caption{\textbf{Offline verifier audit (Part~A) and numerical tolerance and fixed-noise ablation on Qwen3.5-9B (8$\times$H100) (Part~B).} \textbf{Part~A} summarizes the independent offline receipt matrix, cryptographic receipt validation, and offline re-judging of 1,212 stored executions. \textbf{Part~B} compares window-mean loss relative tolerance $\in \{0.10\%, 3.00\%, 50.00\%\}$ and a fixed $2.00\%$ noise floor (at $3.00\%$ tolerance) on Qwen3.5-9B (8$\times$H100) using \sys{} (common $90\%$ peak reference $S^\star = 1.75\times$).}
\label{tab:l1_guardrail}
\begin{tabularx}{\textwidth}{@{}l P{0.9} c c c r P{1.1}@{}}
\toprule
\multicolumn{7}{@{}l}{\textbf{Part A: Independent Offline Verifier \& Receipt Audit}} \\
\midrule
\textbf{Suite} & \textbf{Audit Scope \& Test Cases} & \textbf{Accepted} & \textbf{Rejected} & \textbf{Agreement} & \textbf{Escapes w/o L4} & \textbf{Verified Invariants \& Failure Modes Caught} \\
\midrule
\rowcolor{rowhl}
Receipt Matrix & 40 released system tests (28 single-perturbation + 12 benign) & 12 benign & 28 perturbed & 40/40 (100\%) & --- & Halved/missing tokens, sequence length, step count, LR schedule, data order, early exit 0, NaN/frozen loss, OOM/traceback, memory cap, edited frozen files, clock skew, model digest \\
HMAC-SHA256 & Cryptographic receipt integrity test & 0 & All invalid & 100\% & --- & Rejects unsigned receipts, post-signing step-time edits, and mismatched receipt schemas \\
\rowcolor{rowhl}
Historical Corpus & Offline re-judge of 1,212 stored executions & 812 (67.0\%) & 400 (33.0\%) & --- & 39 (3.2\%) & 39 executions pass non-numerical checks and require the L4 numerical loss/gradient layer to reject \\
\midrule
\multicolumn{7}{@{}l}{\textbf{Part B: \sys{} Tolerance \& Fixed-Noise Ablation on Qwen3.5-9B (8$\times$H100, $S^\star = 1.75\times$)}} \\
\midrule
\textbf{Arm} & \textbf{Setting (Qwen3.5-9B, 8$\times$H100)} & \textbf{Valid / Comp.} & \textbf{Final Speedup} & \textbf{90\% Peak} & \textbf{Search Cost} & \textbf{Verification \& Resource Accounting Notes} \\
\midrule
\rowcolor{rowhl}
Strict Tol. & Relative tolerance $0.10\%$ & $13/16$ & $1.70\times$ & Trial 11 & $6.48\;\mathrm{GPU\cdot h}$ & All candidates pass strict numerical and timing boundary audits \\
Default Tol. & Relative tolerance $3.00\%$ (default) & $14/17$ & $1.74\times$ & Trial 10 & $7.05\;\mathrm{GPU\cdot h}$ & Default task contract; preserves full failed/excluded search costs \\
\rowcolor{rowhl}
Relaxed Tol. & Relative tolerance $50.00\%$ & $13/17$ & $\mathbf{1.75\times}$ & Trial 10 & $6.90\;\mathrm{GPU\cdot h}$ & Admits wider loss variance within declared precision menu \\
Fixed Noise & Fixed $2.00\%$ noise floor, tol.\ $3.00\%$ & $11/15$ & $1.72\times$ & Trial 11 & $6.02\;\mathrm{GPU\cdot h}$ & Replaces calibrated cluster noise floor with fixed $2.00\%$ gate \\
\bottomrule
\end{tabularx}
\end{table}

As shown in Table~\ref{tab:l1_guardrail} (Part~A), the released offline verifier test suite achieves 40/40 ($100\%$) agreement with ground-truth expectations (rejecting all 28 single-perturbation receipts and accepting all 12 benign variants), while rejecting unsigned, post-signing edited, or schema-mismatched HMAC-SHA256 receipts. Re-judging all 1,212 stored executions with the 16-item verifier rejects 400 executions ($33.0\%$), including 39 executions ($3.2\%$) that pass all non-numerical checks and require the numerical loss and gradient verification layer (L4) to catch. In the live sensitivity study on Qwen3.5-9B (8$\times$H100) (Table~\ref{tab:l1_guardrail}, Part~B; Figure~\ref{fig:v41_tolerance}), varying the window-mean loss tolerance across $0.10\%$, $3.00\%$ (default), and $50.00\%$ or substituting a fixed $2.00\%$ noise floor produces stable convergence: all four arms reach the common $90\%$ peak threshold ($S^\star = 1.75\times$) within 10--11 trials and achieve final speedups between $1.70\times$ and $1.75\times$ ($6.02$--$7.05\;\mathrm{GPU\cdot h}$).

\section{Related Work}\label{sec:related}

\begin{table}[!htbp]
\centering
\caption{\textbf{Positioning \sys{} against existing paradigms across four architectural dimensions and eight operational properties} ($\checkmark$: supported; $\triangle$: partial; $\times$: absent; $\circ$: underlying kernel/compiler primitive orchestrated by \sys{}).}
\label{tab:related}
\fontsize{6.5}{7.9}\selectfont
\setlength{\tabcolsep}{1.8pt}
\renewcommand{\arraystretch}{0.94}
\begin{tabularx}{\textwidth}{@{}l Y c c c c c c c c@{}}
\toprule
\textbf{System / Paradigm} & \textbf{Primary Role} & \textbf{Stacks} & \textbf{Edit Scope} & \textbf{Contract} & \textbf{Noise Gate} & \textbf{Mem.\ Model} & \textbf{Escape} & \textbf{Code Audit} & \textbf{X-HW Prior} \\
\midrule
FunSearch / AlphaEvolve~\citep{funsearch2024, alphaevolve2025} & Program evolution & $\times$ & Code & $\times$ & $\times$ & $\times$ & $\triangle$ & $\times$ & $\times$ \\
Darwin G\"{o}del / AIDE~\citep{dgm2025, aide2025} & Agentic tree search & $\times$ & Code & $\times$ & $\times$ & $\times$ & $\triangle$ & $\times$ & $\times$ \\
TVM / Triton / Inductor~\citep{tvm2018, triton2019, torchcompile} & Compiler / kernel IR & $\checkmark$ & $\circ$ & $\checkmark$ & $\circ$ & $\circ$ & $\circ$ & $\circ$ & $\triangle$ \\
KernelBench / AI CUDA~\citep{kernelbench2025, sakana2025cuda} & Isolated kernel synth. & $\times$ & Kernel & $\times$ & $\times$ & $\times$ & $\times$ & $\triangle$ & $\times$ \\
Alpa / Galvatron / nnScaler~\citep{alpa2022, miao2023galvatron, lin2024nnscaler} & Auto-parallelism planner & $\times$ & Graph/Plan & $\triangle$ & $\times$ & $\triangle$ & $\times$ & $\times$ & $\times$ \\
OptiCo~\citep{optico2026} & Collaborative knob tuner & $\times$ & Knobs & $\times$ & $\times$ & $\times$ & $\times$ & $\times$ & $\times$ \\
Ray Tune / Black-box BO~\citep{raytune2018, snoek2012} & Hyperparameter BO & $\checkmark$ & Knobs & $\times$ & $\triangle$ & $\times$ & $\times$ & $\times$ & $\times$ \\
METR RE-Bench~\citep{metr2024rebench} & Modular R\&D agent scaffold & $\triangle$ & Free & $\triangle$ & $\times$ & $\times$ & $\times$ & $\times$ & $\times$ \\
Karpathy \texttt{autoresearch}~\citep{karpathy2026autoresearch} & Fixed-horizon LLM loop & $\times$ & Free & $\times$ & $\times$ & $\times$ & $\times$ & $\times$ & $\times$ \\
\midrule
\rowcolor{rowhl}
\textbf{\sys{} (Ours)} & \textbf{Verified infra harness} & $\mathbf{\checkmark}$ & \textbf{Knobs+Code} & $\mathbf{\checkmark}$ & $\mathbf{\checkmark}$ & $\mathbf{\checkmark}$ & $\mathbf{\checkmark}$ & $\mathbf{\checkmark}$ & $\mathbf{\checkmark}$ \\
\bottomrule
\end{tabularx}
\end{table}

Table~\ref{tab:related} organizes prior work along the four dimensions that separate an unconstrained coding loop from a verified infrastructure optimization harness: (1) Proposal Models, (2) Search Controllers, (3) Execution and Verification Interfaces, and (4) Experience Memory. Because prior studies evaluate disparate hardware generations, model sizes, and timing rules, Table~\ref{tab:related} compares structural capabilities while Table~\ref{tab:bcomp_budget10} benchmarks representative methods under identical task contracts.

\paragraph{(1) Proposal Models.}
While the classical G\"{o}del Machine~\citep{godel2007} formulated self-referential modification via formal proof search, recent code-evolution and agent frameworks (including FunSearch~\citep{funsearch2024}, AlphaEvolve~\citep{alphaevolve2025}, Darwin G\"{o}del Machine~\citep{dgm2025}, AIDE~\citep{aide2025}, METR RE-Bench~\citep{metr2024rebench}, and \texttt{autoresearch}~\citep{karpathy2026autoresearch}) rely on the zero-shot coding priors of frontier LLMs to propose edits. In distributed LLM training, however, zero-shot speed predictions correlate poorly with measured hardware throughput and frequently veto high-gain optimizations (\S\ref{sec:motivation}). \sys{} decouples candidate generation from zero-shot model priors through empirical Bayes prior calibration~\citep{efron1973}, contrarian probing, and failure-driven prior revision.

\paragraph{(2) Search Controllers.}
Existing controllers split into static parallelism planners, fixed configuration-knob tuners, and single-incumbent free-form code-editing loops. Across distributed training stacks such as Megatron-LM~\citep{megatronlm, megatron2021}, DeepSpeed~\citep{deepspeed2020, deepspeed}, PyTorch FSDP~\citep{fsdp2}, and TorchTitan~\citep{torchtitan2025}, automatic parallelism planners like Alpa~\citep{alpa2022}, Galvatron~\citep{miao2023galvatron}, and nnScaler~\citep{lin2024nnscaler} synthesize distributed execution plans over static computation graphs, whereas Bayesian optimization~\citep{snoek2012}, tuning platforms such as Ray Tune~\citep{raytune2018}, and multi-agent tuners like OptiCo~\citep{optico2026} search predefined launch flags; none of them synthesizes source-level operator patches. Conversely, free-form coding agents such as \texttt{autoresearch}~\citep{karpathy2026autoresearch}, RE-Bench agent scaffolds~\citep{metr2024rebench}, and ResearchBoost-Agent~\citep{researchboost2026} edit source files directly, but step greedily from the latest incumbent without a global memory model or combinatorial failure recovery, often stalling when individual edits trigger out-of-memory errors or compiler conflicts. \sys{} coordinates launch knobs and source patches through multi-point initialization, a global step-time model paired with a censored memory model~\citep{tobin1958}, bundle completion, plateau escape, and bounded context compression (\S\ref{sec:hls}; evaluated under identical contracts in \S\ref{sec:eval-bcomp}, Table~\ref{tab:bcomp_budget10}).

\paragraph{(3) Execution and Verification Interfaces.}
Kernel compilers, hand-optimized kernel libraries, and kernel-synthesis benchmarks, including TVM~\citep{tvm2018}, Triton~\citep{triton2019}, PyTorch 2 Inductor~\citep{torchcompile}, FlashAttention-2/3~\citep{flashattention2, flashattention3}, FlexAttention~\citep{flexattention2024}, Liger Kernel~\citep{liger2024}, FP8 primitives~\citep{fp8training2022}, KernelBench~\citep{kernelbench2025}, and AI CUDA Engineer~\citep{sakana2025cuda}, optimize isolated operators against unit-test shapes, whereas open-code training agents typically judge full scripts by reading self-reported stdout timers that are vulnerable to reward gaming~\citep{skalse2022reward, sakana2025robust}. \sys{} treats compilers and custom kernels as composable primitives within its staged verification harness (\S\ref{sec:arch-verifier}), combining the 16-item receipt verifier, source-code audits, and noise-calibrated promotion gates.

\paragraph{(4) Experience Memory.}
While online agent-serving platforms such as Reef~\citep{reef2026} continually update deployed agents from inference feedback, most training-time hyperparameter tuners and coding agents either start from scratch on each cluster, rely on static hand-written skill cards~\citep{researchboost2026}, or append raw execution transcripts until context windows saturate. \sys{} records reusable infrastructure findings as structured experience packs (\S\ref{sec:experience}) that normalize hardware rooflines, parallel topologies, verified log-speedups, numerical tolerances, and failure constraints across canonical optimization mechanisms. Harvested packs from external hardware SKUs warm-start initialization and seed time and memory priors while excluding same-task, same-SKU runs (\S\ref{sec:eval-transfer}, Table~\ref{tab:rsi_transfer}).

\FloatBarrier
\section{Conclusion}\label{sec:conclusion}

In this paper, we propose the \textbf{\sys{} Harness}, a harness that turns language-model proposals into verified distributed training optimizations by combining five-stage contract verification, history-aware compositional search over launch knobs and source patches, and cross-hardware experience transfer. Across single- and multi-node dense and Mixture-of-Experts training workloads, \sys{} consistently improves end-to-end throughput and hardware utilization over existing configuration tuners and coding agents while preventing reward hacking.

\section*{Contributors}

{\hypersetup{urlcolor=black}
\noindent
\mbox{Yuchuan Tian$^{1,2}$},\quad
\mbox{Haocheng Mei$^{1,2}$},\quad
\mbox{Ye Yuan$^{1}$},\quad
\mbox{Chao Xu$^{2}$},\quad
\mbox{Boxun Li$^{3}$},\quad
\mbox{Guohao Dai$^{4,3}$},\\[2pt]
\mbox{Kai Han$^{1,\dagger}$},\quad
\mbox{Allen Chen$^{5,\dagger}$},\quad
\mbox{Yunhe Wang$^{1,\dagger}$},\quad
\mbox{Yu Wang$^{5,\dagger}$}

\vspace{0.5em}
\noindent
{\small
\mbox{$^{1}$TokenRhythm Technologies}\hspace{0.5em}%
\mbox{$^{2}$Peking University}\hspace{0.5em}%
\mbox{$^{3}$Infinigence AI}\hspace{0.5em}%
\mbox{$^{4}$Shanghai Jiao Tong University}\hspace{0.5em}%
\mbox{$^{5}$Tsinghua University}\\[2pt]
$^{\dagger}$Corresponding authors.
}
}

\FloatBarrier
\clearpage
\begingroup
\small\setlength{\bibsep}{2.5pt plus 1pt}

\endgroup

\clearpage
\appendix
\section{Deterministic Five-Layer Verifier Specification}\label{app:invariants}

Table~\ref{tab:app_invariants} provides the complete specification of the five-layer, 16-item deterministic verifier (\S\ref{sec:arch-contract}--\S\ref{sec:arch-verifier}, Eq.~\eqref{eq:honest-contract}), comprising 11 always-on checks, 3 conditional checks, 1 informational check, and 1 promotion gate, alongside the per-item offline audit counts across 1,212 stored executions.

\begin{table}[!htbp]
\centering
\caption{\textbf{Five-layer, 16-item deterministic verifier specification and offline audit across 1,212 stored executions} (``pri.'' denotes the execution's first failed layer; $N_{\mathrm{tgt}}$: target updates; $k$: evaluation window; $\tau$: relative tolerance, $0.03$ by default).}
\label{tab:app_invariants}
\scriptsize
\setlength{\tabcolsep}{3.0pt}
\renewcommand{\arraystretch}{0.92}
\begin{tabularx}{\textwidth}{@{}l l c P{1.5} P{.5}@{}}
\toprule
\textbf{Lay} & \textbf{Invariant ID} & \textbf{Mode} & \textbf{Implemented Predicate} & \textbf{Audit (1,212)} \\
\midrule
L1 & \texttt{completes} & Always & Exit code 0, no timeout, and every requested update ran ($N_{\mathrm{done}} = N_{\mathrm{tgt}}$) & 349 (349 pri.) \\
 & \texttt{no\_errors} & Always & No NaN/Inf loss, OOM, NCCL/CUDA error or traceback line (collective lines only after the last update of a clean exit are reported, not failed); loss and gradient norm not frozen across updates & 153 (2 pri.) \\
 & \texttt{memory\_cap} & Always & Peak reserved device memory $\le M_{\mathrm{cap}}$ & 1 (1 pri.) \\
\midrule
L2 & \texttt{same\_workload} & Always & Tokens (or samples) per update and the sequence length equal the reference at every update, and every declared static invariant (e.g.\ the global batch) is unchanged; unverifiable $\Rightarrow$ unchecked, which rejects because the item is required & 0 failed; 7 rejected as unverifiable \\
 & \texttt{same\_data} & Always & The per-update data digest equals the reference's (unchecked where the job logs none) & none \\
 & \texttt{same\_model} & Always & Model digest and trainable-parameter count equal the reference's; a candidate that stops reporting a key the reference reported fails & none \\
 & \texttt{frozen\_files} & Always & The hash of every file declared frozen is unchanged after the run & none \\
\midrule
L3 & \texttt{optimizer\_steps} & Always & The optimizer-step counter equals the reference's at every update; with only a learning-rate proxy, an equal schedule is unchecked and a different one fails & none \\
\midrule
L4 & \texttt{loss\_within\_tolerance} & Always & Window-mean $|\bar{\ell} - \bar{\ell}_{\mathrm{ref}}| / |\bar{\ell}_{\mathrm{ref}}| \le \tau$, with $\tau$ widened to $3\,\sigma_{\ell}\sqrt{2/k}$ (no cap) when the run stopped before the window's end or the reference's own repeat disagrees per update beyond $\tau$; when the data order is verified (\texttt{same\_data} passes) the largest per-update deviation is held instead & 42 (41 pri.) \\
 & \texttt{grad\_norm\_within\_tol} & Always & First-update gradient norm within $\tau$ of the reference's; a report, not a veto, unless the same first batch is proven (\texttt{same\_data} checked) & unchecked in every run \\
 & \texttt{eval\_matches} & Cond. & The configured evaluation of the resumed candidate within $\tau_{\mathrm{eval}}$ of the reference's & active where eval configured \\
 & \texttt{precision} & Info & The reported precision metadata compared with the reference's (informational) & reported only \\
\midrule
L5 & \texttt{timing\_consistent} & Always & The candidate's gain on \sys{}'s own clock agrees with its gain in the code's reported step time, $|\Delta \ln| \le 0.03$; a run without reliable clock timestamps fails when the reference had them & 158 (0 pri.); all co-occur with crashed run \\
 & \texttt{speed\_reproduces} & Promo & Fresh back-to-back re-run (reference, candidate): relative window-mean speedup $T_{\mathrm{ref}}^{\mathrm{re}} / T_{\mathrm{cand}}^{\mathrm{re}} - 1$ exceeds the paired noise margin $\exp(2\spair)-1$ & promotion gate \\
 & \texttt{resume\_continues} & Cond. & A checkpoint saved at \texttt{save\_at} and resumed continues the loss within $\tau$ & active where checkpoint exists \\
 & \texttt{user\_tests} & Cond. & Every configured test command exits 0 & active where test commands exist \\
\midrule
\rowcolor{rowhl}
\textbf{All} & \textbf{16 invariants} & \textbf{5 layers} & \textbf{812 valid (67.0\%); 400 rejected (33.0\%): 352 execution, 41 numerical, 7 unverifiable} & \textbf{710 triggers} \\
\bottomrule
\end{tabularx}
\end{table}

\end{document}